\documentclass{llncs}
\usepackage{amsmath}
\usepackage{graphicx}
\usepackage{epstopdf}
\usepackage{amssymb}
\usepackage{subfigure}
\usepackage{cite}
\usepackage{algpseudocode}
\usepackage[linesnumbered,ruled]{algorithm2e}
\usepackage{multirow}
\usepackage{booktabs}
\begin{document}
\title{Automatic Liver Segmentation Using an Adversarial Image-to-Image Network}
\vspace{-5mm}
\author{Dong Yang\inst{1}, Daguang Xu\inst{2}, S. Kevin Zhou\inst{2}, Bogdan Georgescu\inst{2}, Mingqing Chen\inst{2}, Sasa Grbic\inst{2}, Dimitris Metaxas\inst{1} \and Dorin Comaniciu\inst{2}}
\institute{Department of Computer Science, Rutgers University, Piscataway, NJ 08854, USA
\and
Medical Imaging Technologies, Siemens Healthcare Technology Center, Princeton, NJ 08540, USA}
\maketitle
\vspace{-8mm}
\begin{abstract}
Automatic liver segmentation in 3D medical images is essential in many clinical applications, such as pathological diagnosis of hepatic diseases, surgical planning, and postoperative assessment. However, it is still a very challenging task due to the complex background, fuzzy boundary, and various appearance of liver. In this paper, we propose an automatic and efficient algorithm to segment liver from 3D CT volumes. A deep image-to-image network (DI2IN) is first deployed to generate the liver segmentation, employing a convolutional encoder-decoder architecture combined with multi-level feature concatenation and deep supervision. Then an adversarial network is utilized during training process to discriminate the output of DI2IN from ground truth, which further boosts the performance of DI2IN. The proposed method is trained on an annotated dataset of 1000 CT volumes with various different scanning protocols (e.g., contrast and non-contrast, various resolution and position) and large variations in populations (e.g., ages and pathology). Our approach outperforms the state-of-the-art solutions in terms of segmentation accuracy and computing efficiency.
\end{abstract}
\vspace{-10mm}
\section{Introduction}
\vspace{-2mm}
\begin{figure}[t]
    \centering
	\vspace{-8mm}
    \centerline{\includegraphics[width=\columnwidth]{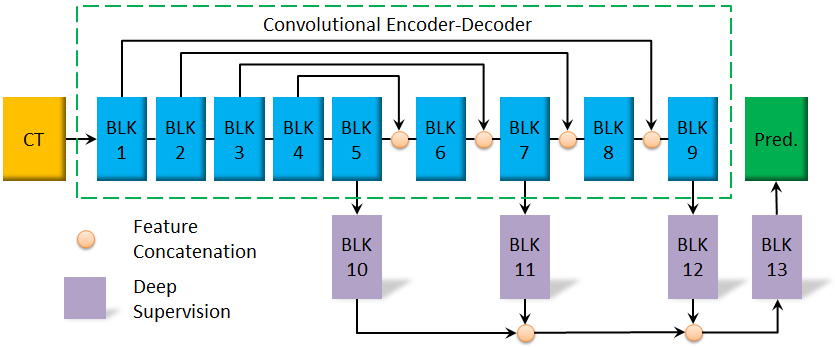}}
	\vspace{-4mm}
    \caption{Proposed deep image-to-image network (DI2IN). The front part is a convolutional encoder-decoder network with feature concatenation, and the backend is deep supervision network through multi-level. Blocks inside DI2IN consist of convolutional and upscaling layers.}
    \vspace{-6mm}
	\label{fig:DI2IN}
\end{figure}
Accurate liver segmentation from three dimensional (3D) medical images
, e.g. computed tomography (CT) or magnetic resonance imaging (MRI)
is essential in many clinical applications, such as pathological diagnosis of hepatic diseases, surgical planning, and postoperative assessment. However, automatic liver segmentation is still a highly challenging task due to the complex background, fuzzy boundary, and various appearance of liver in medical images.

To date, several methods have been proposed for automatic liver segmentation from 3D CT scans. Generally, they can be categorized into non-learning-based and learning-based approaches. Non-learning-based approaches usually rely on the statistical distribution of the intensity, including atlas-based \cite{Linguraru2009Atlas}, active shape model (ASM)-based \cite{Kainmuller2007ASM}, levelset-based \cite{Lee2007Levelset}, and graph-cut-based \cite{Massoptier2007Graphcut} methods, etc.
On the other hand, learning-based approaches take the advantage of hand-crafted features to train the classifiers to achieve good segmentation. For example, in \cite{Ling2008MSL}, the proposed hierarchical framework applies marginal space learning with steerable features to handle the complicated texture pattern near the liver boundary.

Until recently, deep learning has been shown to achieve superior performance in various challenging tasks, such as classification, segmentation, and detection. Several automatic liver segmentation approaches based on convolutional neural network (CNN) have been proposed. Dou, et, al. \cite{Dou2016Liver} demonstrated a fully convolutional network (FCN) with deep supervision, which can perform end-to-end learning and inference. The output of FCN is refined with a fully connected conditional random field (CRF) approach. Similarly, Christ, et, al. \cite{Christ2016Liver} proposed cascaded FCNs followed by CRF refinement. Lu, et, al. \cite{Lu2017Liver} used a FCN with graph-cut based refinement. Although these methods demonstrated good performance, they all used pre-defined refinement approaches. For example, both CRF and graph-cut methods are limited to the use of pairwise models, and time-consuming as well. They may cause serious leakage at boundary regions with low contrast, which is common in liver segmentation.

Meanwhile, Generative Adversarial Network (GAN) \cite{Goodfellow2014GAN} has emerged as a powerful framework in various tasks. It consists of two parts: generator and discriminator. The generator tries to produce the output that is close to the real samples, while the discriminator attempts to distinguish between real and generated samples.
Inspired by \cite{Luc2016GAN}, we propose an automatic liver segmentation approach using an adversarial image-to-image network (DI2IN-AN). A deep image-to-image network (DI2IN) is served as the generator to produce the liver segmentation. It employs a convolutional encoder-decoder architecture combined  with  multi-level  feature  concatenation  and  deep  supervision.  Our network tries to optimize a conventional multi-class cross-entropy loss together with an adversarial term that aims to distinguish between the output of DI2IN and ground truth.
Ideally, the discriminator pushes the generator's output towards the distribution of ground truth, so that it has the potential to enhance generator's performance by refining its output. Since the discriminator is usually a CNN which takes the joint configuration of many input variables, it embeds the higher-order potentials into the network (the geometric difference between prediction and ground truth is represented by the trainable network model instead of heuristic hints). The proposed method also achieves higher computing efficiency since the discriminator does not need to be executed at inference.

All previous liver segmentation approaches were trained using dozens of volumes which did not take the full advantage of CNN. In contrast, our network leverages the knowledge of an annotated dataset of 1000+ CT volumes with various different scanning protocols (e.g., contrast and non-contrast, various resolution and position) and large variations in populations (e.g., ages and pathology). To the best of our knowledge, our experiment is the first time that more than 1000 annotated 3D CT volumes are adopted in liver segmentation tasks. The experimental result shows that training with such a large dataset significantly improves the performance and enhances the robustness of the network.

\section{Methodology}
\subsection{Deep Image-to-Image Network (DI2IN) for Liver Segmentation}
In this section, we present a deep image-to-image network (DI2IN), which is a multi-layer convolutional neural network (CNN), for the liver segmentation.
The segmentation task is defined as the voxel-wise binary classification.

DI2IN takes the entire 3D CT volumes as input, and outputs the probability maps that indicate how likely voxels belongs to the liver region.
As shown in Fig. \ref{fig:DI2IN}, the main structure of DI2IN is designed following a symmetric way as a convolutional encoder-decoder.
All blocks in DI2IN consist of 3D convolutional and bilinear upscaling layers. The details of the network is described in Fig. \ref{fig:TABLE}.

In the encoder part of DI2IN, only the convolution layers are used in all blocks.
In order to increase the receptive field of neurons and lower the GPU memory consumption, we set stride as 2 at some layers and reduce the size of feature maps.
Moreover, larger receptive field covers more contextual information and helps to preserve liver shape information in the prediction.
The decoder of DI2IN consists of convolutional and bilinear upscaling layers.
To enble end-to-end prediction and training, the upscaling layers are implemented as bilinear interpolation to enlarge the activation maps.
All convolutional kernels are $3\times3\times3$.
The upscaling factor in decoder is 2 for $x,y,z$ dimension.
The Leaky rectified linear unit (Leaky ReLU) and batch normalization are adopted in all convolutional layers for proper gradient back-propagation.

In order to further improve the performance of DI2IN, we adopt several mainstream technologies with the necessary changes \cite{ronneberger2015u, merkow2016dense, Dou2016Liver}.
First, we use the feature layer concatenation in DI2IN.
Fast bridges are built directly from the encoder layers to the decoder layers.
The bridges pass the information from the encoder forward and then concatenate it with the decoder feature layers.
The combined feature is used as the input for the next convolution layer.
Following the steps above to explicitly combine advanced and low-level features, DI2IN benefits from local and global contextual information.
The deep supervision of the neural network during end-to-end training is shown to achieve good boundary detection and segmentation results.
In the network, we introduced a more complex deep supervision scheme to improve performance.
Several branches are separated from layers of the decoder section of main DI2IN.
With the appropriate upscaling and convolution operations, the output size of each channel for all branches matches the size of the input image (Upscaling factors are 16,4,1 in block 10,11,12 repectively).
By calculating the loss item $ l_ {i} $ with the same ground truth data, the supervision is enforced at the end of each branch $i$.
In order to further utilize the results of different branches, the final output is determined by the convolution operations of all branches with the leaky ReLU.
During training, we apply binary cross entropy loss to each voxel of the output layers.
The total loss $ l_ {total} $ is the weighted combination of loss terms for all output layers, including the final output layer and the output layers for all branches, as follows:
\[l_{total} = \sum_{i}w_{i}\cdot l_{i} + w_{final}\cdot l_{final}\]
\vspace{-8mm}
\subsection{Network Improvement with Adversarial Training}
\begin{figure}[t]
    \centering
    \vspace{-5mm}
    \centerline{\includegraphics[width=0.9\columnwidth]{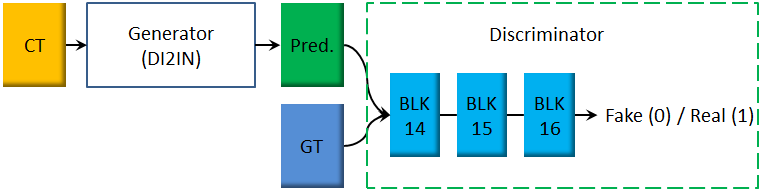}}
	\vspace{-4mm}
    \caption{Proposed adversarial training scheme. The generator produces the segmentation prediction, and discriminator classifies the prediction and ground truth during training.}
    \vspace{-4mm}
	\label{fig:GAN}
\end{figure}

\begin{figure}[t]
    \centering
	\vspace{-8mm}
    \centerline{\includegraphics[width=\columnwidth]{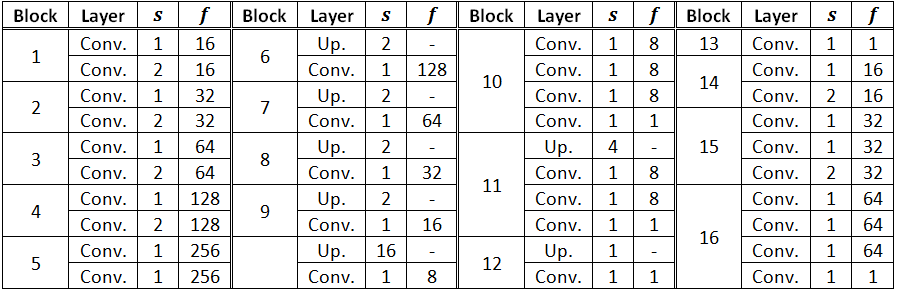}}
	\vspace{-4mm}
    \caption{Parametric setting of blocks in neural network. $s$ stands for the stride, $f$ is filter number. \textit{Conv.} is convolution, and \textit{Up.} is bilinear upscaling.}
    \vspace{-2mm}
	\label{fig:TABLE}
	\vspace{-4mm}
\end{figure}

We adopt the prevailing idea of the generative adversarial networks to boost the performance of DI2IN.
The proposed scheme is shown in Fig.\ref{fig:GAN}.
An adversarial network is adopted to capture the high-order appearance information, which distinguishes between the ground truth and the output from DI2IN.
In order to guide the generator to better prediction, the adversarial network provides an extra loss function for updating the parameters of generator during training.
The purpose of the extra loss is to make the prediction as close as possible to the ground truth labeling.
We adopt the binary cross-entropy loss for training of the adversarial network.
$D$ and $G$ represent the discriminator and generator (DI2IN, in the context), respectively.
For the discriminator $D\left(Y;\theta^D\right)$, the ground truth label $Y_{gt}$ is assigned as one, and the prediction $Y_{pred}=G\left(X;\theta^G\right)$ is assigned as zero where $X$ is the input CT volumes.
The structure of discriminator network $D$ is shown in Fig. \ref{fig:TABLE}.
The following objective function is used in training the adversarial network:
\begin{equation}
\begin{split}
l_{D} & =-\mathbb{E}_{y\sim p_{gt}}\log \left(D\left(y;\theta^D\right)\right)-\mathbb{E}_{y'\sim p_{pred}}\log \left(1-D\left(y';\theta^D\right)\right)\\
 & =-\mathbb{E}_{y\sim p_{gt}}\log \left(D\left(y;\theta^D\right)\right)-\mathbb{E}_{x\sim p_{data}}\log \left(1-D\left(G\left(x;\theta^G\right);\theta^D\right)\right)
\end{split}
\label{eq:adv}
\end{equation}
During the training of network $D$, the gradient of loss $l_D$ is propagated back to update the parameters of the generator network (DI2IN).
At this stage, the loss for $G$ has two components shown in the Equation \ref{eq:seg1}.
The first component is the conventional segmentation loss $l_b$: voxel-wise binary cross-entropy between the prediction and ground truth.
Minimizing the second loss component enables the discriminator $D$ to confuse the ground truth with the prediction from $G$.
\begin{equation}
\begin{split}
l_{G} & = \mathbb{E}_{y\sim p_{pred},y'\sim p_{gt}}\left[l_{seg}\left(y,y'\right )\right]-\lambda\mathbb{E}_{y\sim p_{pred}}\log \left(1-D\left (y;\theta^D\right)\right)\\
& = \mathbb{E}_{y\sim p_{pred},y'\sim p_{gt}}\left[l_{seg}\left(y,y'\right )\right]-\lambda\mathbb{E}_{x\sim p_{data}}\log \left(1-D\left(G\left(x;\theta^G\right);\theta^D\right)\right)
\end{split}
\label{eq:seg1}
\end{equation}
Following suggestions in \cite{Goodfellow2014GAN}, we replace $-\log \left(1-D\left(G\left(x\right)\right)\right)$ with $\log \left(D\left(G\left ( X \right )\right)\right)$.
In another word, we would like to maximize the probability that prediction to be the ground truth in Equation \ref{eq:seg1}, instead of minimizing the probability that prediction not to be the generated label map.
Such replacement provides strong gradient during training of $G$ and speed up the training process in practice.
\begin{equation}
l_{G} = \mathbb{E}_{y\sim p_{pred},y'\sim p_{gt}}\left[l_{seg}\left(y,y'\right )\right]+\lambda\mathbb{E}_{x\sim p_{data}}\log D\left(G\left(x;\theta^G\right);\theta^D\right)
\label{eq:seg2}
\end{equation}
The generator and discriminator are trained alternatively for several times shown in Algorithm \ref{algo:GAN}, until the discriminator is not able to easily distinguish between ground truth label and the output of DI2IN.
After the training process, the adversarial network is no longer required at inference.
The generator itself can provide high quality segmentation results and its performance is improved.
\begin{algorithm}[t]
 \SetKwInOut{Input}{Input}
 \SetKwInOut{Output}{Output}
 \Input{pre-trained generator (DI2IN) with weights $\theta^G_0$}
 \Output{updated generator weights $\theta^G_1$}
  \For{number of training iterations}{
  \For{$k_D$ steps}{
    sample a mini-batch of training images $x\sim p_{data}$\;
    generate prediction $y_{pred}$ for $x$ with $G\left(x;\theta^G_0\right)$\;
    $\theta^D\leftarrow$ propagate back the stochastic gradient $\triangledown l_D\left(y_{gt},y_{pred}\right)$\;
  }
  \For{$k_G$ steps}{
    sample a mini-batch of training images $x'\sim p_{data}$\;
    generate $y'_{pred}$ for $x'$ with $G\left(x';\theta^G_0\right)$ and compute $D\left(G\left(x'\right)\right)$\;
    $\theta^G_1\leftarrow$ propagate back the stochastic gradient $\triangledown l_G\left(y'_{gt},y'_{pred}\right)$\;
  }
  $\theta^G_0\leftarrow\theta^G_1$
 }
 \caption{Adversarial training of generator and discriminator.}
 \label{algo:GAN}
\end{algorithm}
\vspace{-3mm}
\section{Experiments}
\vspace{-3mm}
Most public dataset for liver segmentation only consists of tens of cases. For example, the MICCAI-SLiver07 \cite{Sliver2009} dataset only contains 20 CT volumes for training and 10 CT volumes for testing. All the data are contrast enhanced. Such a small dataset is not suitable to show the power of CNN: it has been well known that neural network trained with more labelled data can usually achieve much better performance. Thus, in this paper, we collected more than 1000 CT volumes. The liver of each volume was delineated by human experts. These data covers large variations in populations, contrast phases, scanning ranges, pathologies, and field of view (FOV), etc. The inter-slice distance varies from 0.5mm to 7.0mm. All scans covers the abdominal regions but may extend to head and feet. Tumor can be found in multiple cases. The volumes may also have various other disease. For example, pleural effusion, which brights the lung region and changes the pattern of upper boundary of liver.
Then we collected additional 50 volumes from clinical sites for the independent testing.
The livers of these data were also annotated by human experts for the purpose of evaluation.
We down-sampled the dataset into 3.0mm resolution isotropically to speed up the processing and lower the consumption of computing memory without loss of accuracy.
Training DI2IN from scratch takes 200 iterations using stochastic gradient descent with a batch size of 4 samples.
The learning rate is 0.01 in the beginning and divided by 10 after 100 iterations.
In the adversarial training, we set $\lambda$ to 0.01, and the number of overall training iterations is 100.
For training $D$, $k_D$ is 10 and the mini-batch size is 8.
For training $G$, $k_G$ is 1 and the mini-batch size is 4.
Less training iterations are required for $G$ than that for $D$ because $G$ is pre-trained before adversarial training.
$w_i$ is set as 1 in the loss.

\begin{figure}[htb]
    \centering
	\vspace{-0mm}
    \centerline{\includegraphics[width=\columnwidth]{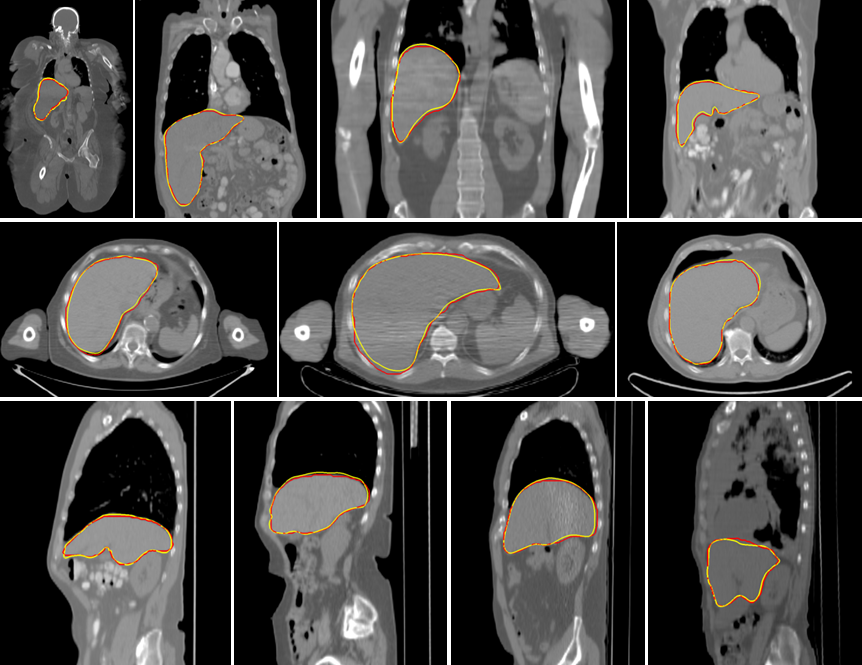}}
	\vspace{-4mm}
    \caption{Visual Results from different views. Yellow meshes are ground truth. Red ones are the prediction from DI2IN-AN.}
    \vspace{-4mm}
	\label{fig:RESULT}
	\vspace{-2mm}
\end{figure}

Table 1 compares the performance of five different methods. The first method, the hierarchical, learning-based algorithm proposed in \cite{Ling2008MSL}, was trained using 400 CT volumes. More training data did not show performance improvement for this method.
For comparison purpose, the DI2IN network, which is similar to deep learning based algorithms proposed in \cite{Dou2016Liver,Christ2016Liver,Lu2017Liver} without post-processing steps, and the DI2IN-AN were trained using the same 400 cases.
Both the DI2IN network and the DI2IN-AN were also trained using all 1000+ CT volumes. The average symmetric surface distance (ASD) and dice coefficients are computed for all methods on the test data. As shown in Table 1, DI2IN-AN achieves the best performance in both evaluation metrics. All deep learning based algorithms outperform the classic learning based algorithm with the hand-craft features, which shows the power of CNN. The results show that more training data enhances the performance of both DI2IN and DI2IN-AN. Take DI2IN for example, training with 1000+ labelled data improves the mean ASD by 0.23mm and the max ASD by 3.84mm compared to training with 400 labelled data.
Table 1 also shows that the adversarial structure can further boost the performance of DI2IN. The maximum ASD error is also reduced. Typical test samples are provided in Fig. \ref{fig:RESULT}.
We also tried CRF and graph cut to refine the output of DI2IN. However, the results became worse, since a large portion of testing data had no contrast and the boundary of liver bottom at many locations was very fuzzy. CRF and graph cut both suffer from serious leakage in these situations. Using an NVIDIA TITAN X GPU and the Theano/Lasagne library, the run time of our algorithm is less than one second, which is significantly faster than most of the current approaches. For example, it requires 1.5 minutes for one case in \cite{Dou2016Liver}. More experimental results can be found in the supplementary material.

Our proposed DI2IN has clear advantages comparing with other prevailing methods. First, previous studies show that DI2IN, which incorporates the encoder-decoder structure, skip connections, and deep supervision scheme within one framework, has better structure design than U-Net or deep supervised network (DSN) for 3D volumetric datasets \cite{Dou2016Liver,merkow2016dense}. DI2IN is a different design from other prevailing networks, but it gathers the merits of them. Second, the CNN-based methods (no upsampling or deconvolution) are often time-consuming at inference, and their performance is sensitive to the selection of training sample. We examined the aforementioned networks with internal implementation, and DI2IN achieved better performance (20\% improvement in terms of average symmetric surface distance).
\begin{table}[t]
\renewcommand{\arraystretch}{1.3}
\renewcommand{\multirowsetup}{\centering}
\setlength{\belowrulesep}{0pt}
\setlength{\aboverulesep}{0pt}
\caption{Comparison of five methods on 50 unseen CT data.}
\begin{center}
\begin{tabular}{|c|c|c|c|c|c|c|c|c|}
\hline
\multirow{2}{*}{Method} & \multicolumn{4}{|c|}{ASD (mm)} & \multicolumn{4}{|c|}{Dice}\\
\cmidrule{2-9}
& Mean & Std & Max & Median & Mean & Std & Min & Median\\
\cmidrule{1-9}
Ling \emph{et al.} (400) \cite{Ling2008MSL} & 2.89 & 5.10 & 37.63 & 2.01 & 0.92 & 0.11 & 0.20 & 0.95 \\
\cmidrule{1-9}
DI2IN (400) & 2.25  & 1.28 & 10.06 & 2.0 & 0.94 & 0.03 & 0.79 & 0.94\\
\cmidrule{1-9}
DI2IN-AN (400) & 2.00 & 0.95 & 7.82 & 1.80 & 0.94 & 0.02 & 0.85 & 0.95\\
\cmidrule{1-9}
DI2IN (1000) & 2.15 & 0.81 & 6.51 & 1.95 & 0.94 & 0.02 & 0.87 & 0.95 \\
\cmidrule{1-9}
DI2IN-AN (1000) & \textbf{1.90} & 0.74 & \textbf{6.32} & \textbf{1.74} & \textbf{0.95} & 0.02 & \textbf{0.88} & 0.95\\
\hline
\end{tabular}
\end{center}
\vspace{-8mm}
\end{table}
\vspace{-3mm}
\section{Conclusion}
In this paper, we proposed an automatic liver segmentation algorithm based on an adversarial image-to-image network.
Our method achieves good segmentation quality as well as faster processing speed. The network is trained on an annotated dataset of 1000+ 3D CT volumes. We demonstrate that training with such a large dataset can improve the performance of CNN by a large margin.

 
%
\end{document}